\documentclass[sigconf]{acmart}%

\usepackage{natbib} 
\usepackage{booktabs}
\usepackage{comment}
\usepackage{graphicx}
\usepackage{subfig}
\usepackage{fancyhdr}
\usepackage{amsmath}
\usepackage{xurl}
\usepackage{adjustbox}
\usepackage{arydshln}
\usepackage{relsize}
\usepackage{wasysym}
\usepackage{multirow}
\usepackage[makeroom]{cancel}
\usepackage{scalerel,graphicx,xparse}
\usepackage[capitalize,nameinlink]{cleveref}
\usepackage{algorithm}
\usepackage{algpseudocode}
\usepackage{algorithm}
\usepackage{lscape}
\usepackage{graphicx}
\usepackage{enumitem}

\graphicspath{{../}{images/}}
\usepackage{hyperref}

\definecolor{navy}{rgb}{0.1, 0.1, 0.8}
\definecolor{gray}{rgb}{0.4, 0.4, 0.4}
\definecolor{olive}{rgb}{0.1, 0.5, 0.1}
\definecolor{ruby}{rgb}{0.8, 0.1, 0.3}
\definecolor{darkpastelgreen}{rgb}{0.01, 0.75, 0.24}
\definecolor{celestialblue}{rgb}{0.29, 0.59, 0.82}
\definecolor{coral}{rgb}{1.0, 0.5, 0.31}
\definecolor{blue}{rgb}{0.23, 0.44, 0.62}
\definecolor{Goldenrod}{rgb}{0.8,0.8,0}
\definecolor{pinky}{RGB}{255,20,147}  

\usepackage{soul}
\usepackage[colorinlistoftodos,prependcaption,textsize=tiny,textwidth=15mm]{todonotes}
\usepackage{xspace}

\newcommand{\eat}[1]{}

\NewDocumentCommand\iclogo{}{\raisebox{0.2\height}{\scalerel*{\includegraphics[height=5em]{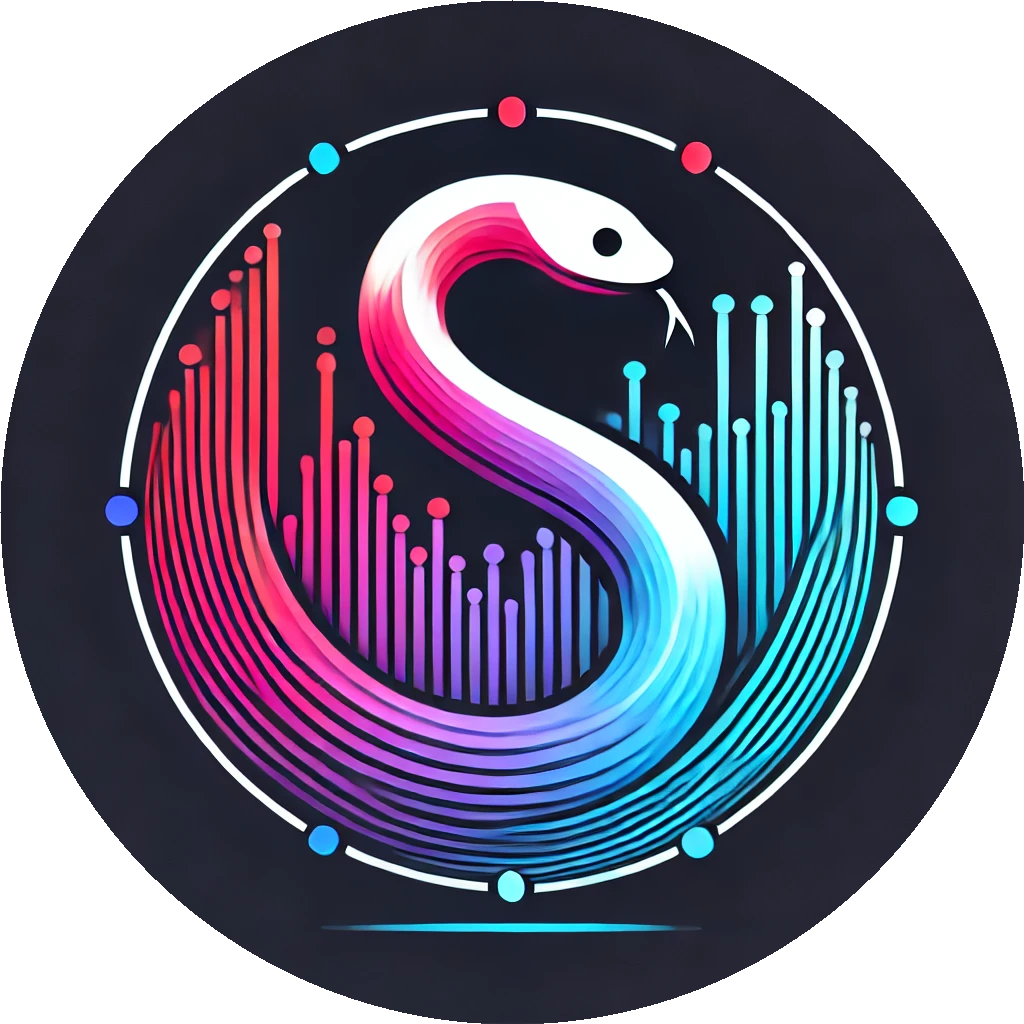}}{50pt}}\xspace}
\newcommand{\icmamba}{IC-Mamba\iclogo}

\AtBeginDocument{%
  }

\copyrightyear{2025}
\acmYear{2025}
\setcopyright{cc}
\setcctype{by}
\acmConference[WWW '25]{Proceedings of the ACM Web Conference 2025}{April 28-May 2, 2025}{Sydney, NSW, Australia}
\acmBooktitle{Proceedings of the ACM Web Conference 2025 (WWW '25), April 28-May 2, 2025, Sydney, NSW, Australia}
\acmDOI{10.1145/3696410.3714527}
\acmISBN{979-8-4007-1274-6/25/04}

\begin{document}


\title{Before It's Too Late: A State Space Model for the Early Prediction of Misinformation and Disinformation Engagement}
\author{Lin Tian}
\affiliation{%
  \institution{University of Technology Sydney}
  \city{Sydney}
  \country{Australia}}
\email{lin.tian-3@uts.edu.au}

\author{Emily Booth}
\affiliation{%
  \institution{University of Technology Sydney}
  \city{Sydney}
  \country{Australia}}
\email{emily.booth@uts.edu.au}

\author{Francesco Bailo}
\affiliation{%
  \institution{The University of Sydney}
  \city{Sydney}
  \country{Australia}}
\email{francesco.bailo@sydney.edu.au}

\author{Julian Droogan}
\affiliation{%
  \institution{Macquarie University}
  \city{Sydney}
  \country{Australia}}
\email{julian.droogan@mq.edu.au}

\author{Marian-Andrei Rizoiu}
\affiliation{%
  \institution{University of Technology Sydney}
  \city{Sydney}
  \country{Australia}}
\email{marian-andrei.rizoiu@uts.edu.au}


\begin{abstract}
In today’s digital age, conspiracies and information campaigns can emerge rapidly and erode social and democratic cohesion.
While recent deep learning approaches have made progress in modeling engagement through language and propagation models, they struggle with irregularly sampled data and early trajectory assessment. We present \icmamba, a novel state space model that forecasts social media engagement by modeling interval-censored data with integrated temporal embeddings. 
Our model excels at predicting engagement patterns within the crucial first 15-30 minutes of posting (RMSE 0.118-0.143), enabling rapid assessment of content reach. 
By incorporating interval-censored modeling into the state space framework, IC-Mamba captures fine-grained temporal dynamics of engagement growth, achieving a 4.72\% improvement over state-of-the-art across multiple engagement metrics (likes, shares, comments, and emojis).
Our experiments demonstrate IC-Mamba's effectiveness in forecasting both post-level dynamics and broader narrative patterns (F1 0.508-0.751 for narrative-level predictions). 
The model maintains strong predictive performance across extended time horizons, successfully forecasting opinion-level engagement up to 28 days ahead using observation windows of 3-10 days. 
These capabilities enable earlier identification of potentially problematic content, providing crucial lead time for designing and implementing countermeasures.
Code is available at: \url{https://github.com/ltian678/ic-mamba}.
An interactive dashboard demonstrating our results is available at: \url{https://ic-mamba.behavioral-ds.science/}.
\end{abstract}




\maketitle

\section{Introduction}

On 28 October 2017, an anonymous 4chan user made a brief yet impactful post on the platform claiming that Hillary Clinton was to be arrested in the coming days\footnote{\url{https://www.bellingcat.com/news/americas/2021/01/07/the-making-of-qanon-a-crowdsourced-conspiracy/}}.
On 6 January 2021, devotees of then-outgoing President Donald Trump stormed the United States Capitol building in an act of domestic terrorism designed to prevent President-elect Joe Biden's election victory from being confirmed. Five people died during and in the immediate aftermath of the attack, and an additional four died in the subsequent months\footnote{\url{https://www.factcheck.org/2021/11/how-many-died-as-a-result-of-capitol-riot/}}; and over 140 police officers were injured\footnote{\url{https://www.nytimes.com/2021/08/03/us/politics/capitol-riot-officers-honored.htm}}. 
Investigations by the Associated Press of the online social media profiles of over 120 of the rioters revealed high levels of adherence to the QAnon conspiracy theory that had begun just four years prior on 4chan\footnote{\url{https://apnews.com/article/us-capitol-trump-supporters-1806ea8dc15a2c04f2a68acd6b55cace}}.
This incident highlights how social media platforms can accelerate the spread of harmful content, particularly misinformation (false information shared without intent to harm) and disinformation (deliberately created and shared false information)~\citep{lazer2018science,scheufele2019science}.

\begin{figure}[t]
  \centering
  \includegraphics[width=\linewidth]{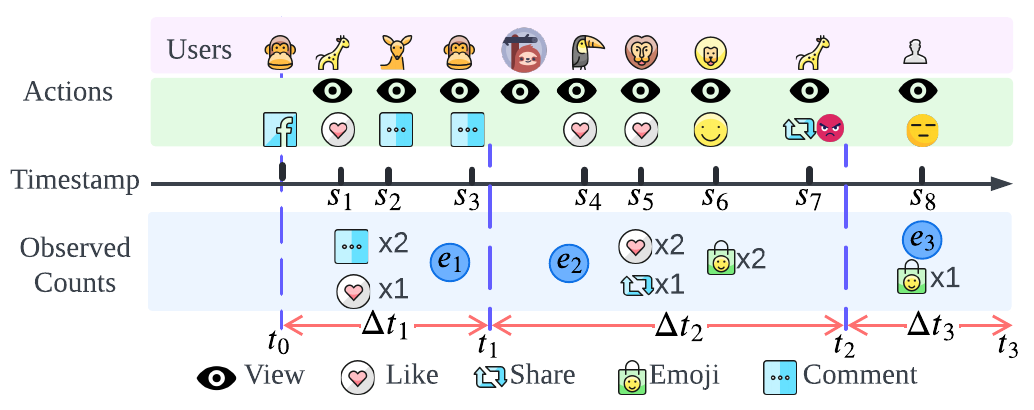}
  \caption{
    Illustration of interval-censored social media engagement data. Following a post's creation at $t_0$, users perform engagement actions (view, like, comment, share, emoji) at timestamps $s_1$ through $s_8$. While individual actions occur continuously, engagement data is only collected at discrete observation points $t_j$, where each engagement vector $e_j$ captures the cumulative counts of different interaction types over intervals of length $\Delta t_j = t_{j+1} - t_j$.
  }
  \label{fig:sample_ic_mamba}
  \Description{social engagement matters sample}
\end{figure}

Given these ongoing impacts, the question must be asked: what if we could have seen this coming? More specifically, what if it had been possible to forecast user engagement with fringe ideologies before they morph into widespread movements? 
We introduce \icmamba, a model capable of forecasting user engagement with online content. We go beyond the level of atomic posts to forecast the number of likes, shares, emoji reactions, and comments for ``emerging opinions'' -- particular worldviews supported across a series of posts.
Our framework can forecast the arrival rate of posts supporting an opinion, and forecast the engagement for each, obtaining estimates of the total level of engagement for the entire opinion.
Our analysis leverages CrowdTangle~\cite{crowdtangle} data with interval-censored engagement metrics, where observations are made at discrete time points with engagement counts recorded for each interval (as seen in \cref{fig:sample_ic_mamba}).

Recent deep learning approaches have made progress in modeling social media engagement through different architectural innovations: language models to capture coordinated posting behaviors~\citep{atanasov2019predicting,tian2021rumour,tian2022duck,tian2023metatroll} and propagation models to model information
diffusion~\citep{zannettou2019disinformation,im2020still,luceri2024unmasking,kong2023interval,Kong2021,Kong2020,Kong2020a,Zhang2019,Kong2018}. 
State space models have also demonstrated strong performance on sequential prediction tasks ~\citep{mamba,mamba2}, with their latent state representations theoretically well suited for temporal dependencies. 
However, these approaches face two key limitations when applied to mis/disinformation engagement forecasting: 
(1) they primarily focus on classification tasks rather than quantifying future temporal patterns of engagement, and 
(2) they struggle with the irregularly sampled nature of viral content.

\noindent\textbf{Our main contributions} address three research questions (RQs) at the intersection of temporal modeling and social media dynamics:

RQ1: \textit{How can we effectively model irregular temporal patterns in social media engagement?}: Through IC-Mamba's time-aware embeddings and state space model architecture, we capture the dynamics of online interactions, achieving a 4.72\% improvement over the state-of-the-art approaches. 

RQ2:\textit{Can we predict viral potential within the critical early window? }: \icmamba shows strong performance in the crucial 15-30 minute post-publication window (RMSE 0.118-0.143), while capturing both granular post-level dynamics and broader narrative patterns (F1 0.508-0.751 for narrative-level predictions).

RQ3: \textit{How can we forecast engagement with emerging opinions early? Can we improve the accuracy and confidence of these forecasts as engagement data streams in over time?}: Our experiments show the model effectively forecasts engagement dynamics early, using 3-, 7-, and 10-day windows to predict spreading patterns up to 28 days, with performance improving as more data streams in. 

As a tool, \icmamba streamlines the work of human experts, enabling earlier identification of problematic content, and therefore, providing more time to design and implement countermeasures.

 \begin{table}[t]
     \caption{
         Information used in related studies and our work (\icmamba). V: views, S: shares/retweets,
         C: comments, L: likes, E: emojis, WB: Weibo, YT: Youtube, FB: Facebook.
     }
     \label{tab:related_work}
     \begin{tabular}{lcccccc}
         \toprule
                                               & V          & S          & C          & L          & E          & platform  \\ \toprule
         DeepCas~\citep{li2017deepcas}         &            & \checkmark &            &            &            & X         \\
         DeepHawkes~\citep{cao2017deephawkes}  &            & \checkmark &            &            &            & X         \\
         Topo-LSTM~\citep{wang2017topological} &            & \checkmark &            &            &            & X         \\
         HIP ~\citep{rizoiu2017expecting}      & \checkmark & \checkmark &            &            &            & YT        \\
         DeepInf~\citep{qiu2018deepinf}        &            & \checkmark &            & \checkmark &            & WB and X  \\
         B-Views~\cite{wu2018beyond}           & \checkmark &            &            &            &            & YT        \\
         SNPP ~\citep{ding2019social}          &            &            &            & \checkmark &            & X         \\
         CasFlow~\citep{xu2021casflow}         &            & \checkmark &            &            &            & WB        \\
         MBPP~\cite{rizoiu2022interval}        & \checkmark & \checkmark &            &            &            & YT        \\
         IC-TH~\cite{kong2023interval}         &            & \checkmark &            &            &            & X         \\
         OMM~\citep{calderon2024opinion}       &            & \checkmark &            &            &            & X,YT\& FB \\
         \midrule
         \icmamba                              &            & \checkmark & \checkmark & \checkmark & \checkmark & FB        \\
         \bottomrule
     \end{tabular}
 \end{table}
\section{Related Work}
This section reviews relevant literature in two key areas that underpin our approach to modeling and predicting social engagements during outbreak events such as information operations and natural disasters like the infamous 2019-2020 Australian Bushfires:
popularity and engagement prediction on social media platforms, and
state space models for sequence modeling and prediction.

\subsection{Popularity and Engagement Prediction}
Social media engagement prediction research spans various platforms and prediction tasks. 
DeepCas~\citep{li2017deepcas} used random walk and attention mechanisms to predict final cascade size on X, while SNPP~\citep{ding2019social} applied temporal point process with a gated recurrent unit architecture for tweet repost count prediction.
Topo-LSTM~\citep{wang2017topological} incorporated user interaction sequences through a topological structure for retweet prediction, while DeepInf~\citep{qiu2018deepinf} on Weibo and X predicted user retweets, likes, and following behaviors using graph convolutional networks within the 2-hop neighborhood. 
CasFlow~\citep{xu2021casflow} used hierarchical attention networks for modeling Weibo reposts.
Several approaches have focused on handling temporal dynamics in engagement prediction. DeepHawkes~\citep{cao2017deephawkes} integrated reinforcement learning with Hawkes processes for retweet cascade prediction.
For YouTube, HIP~\citep{rizoiu2017expecting} and MBPP~\citep{rizoiu2022interval,Calderon2025} advanced temporal modeling for views prediction, with B-Views~\citep{wu2018beyond} specifically addressing cold-start scenarios.
Recently, IC-TH~\citep{kong2023interval} tackled the challenge of incomplete observations in retweet prediction on X, while OMM~\citep{calderon2024opinion} and BMH~\cite{Calderon2024b} proposed a mathematical framework for shares prediction on X, YouTube, and Facebook.
While these approaches have advanced cascade modeling, existing interval-censored methods (IC-TH, MBPP) focus on single-post dynamics without considering broader opinion-level patterns. Our work extends beyond individual post predictions to model collective opinion engagement on Facebook, and we include these interval-censored models along with neural approaches (TH~\cite{zuo2020transformer}) as baselines.



\subsection{State Space Model in Sequence Modeling}
State Space Models (SSMs) have recently emerged as a robust alternative to traditional sequence modeling approaches, particularly for long-range dependency capture~\citep{hasani2021liquid,rangapuram2018deep}. Subsequent work has showed their efficiency in processing extremely long sequences~\citep{gu2022efficiently,dao2022flashattention} and competitive performance in language modeling~\citep{mamba,mamba2}.
Despite these advancements, few studies have applied modern SSM architectures to predict or analyze social media engagement, particularly in the context of disinformation campaigns or crisis events. 
Most existing methods rely on graph-based~\citep{lu2023continuous}, RNN~\citep{wang2017cascade}, or transformer approaches~\citep{zuo2020transformer} that typically assume uniform sampling or discret snapshots. Such assumptions often overlook fine-grained temporal patterns crucial to disinformation campaigns or crisis events.
In contrast, modern SSMs can naturally handle non-uniform intervals and continuous-time dynamics, making them well-suited for rapidly unfolding social media processes. 
Our work bridges this gap by extending the Mamba architecture to handle non-uniform intervals, while identifying misinformation opinions and disinformation narratives.


\begin{figure*}[tbp]
  \centering
  \includegraphics[width=\linewidth]{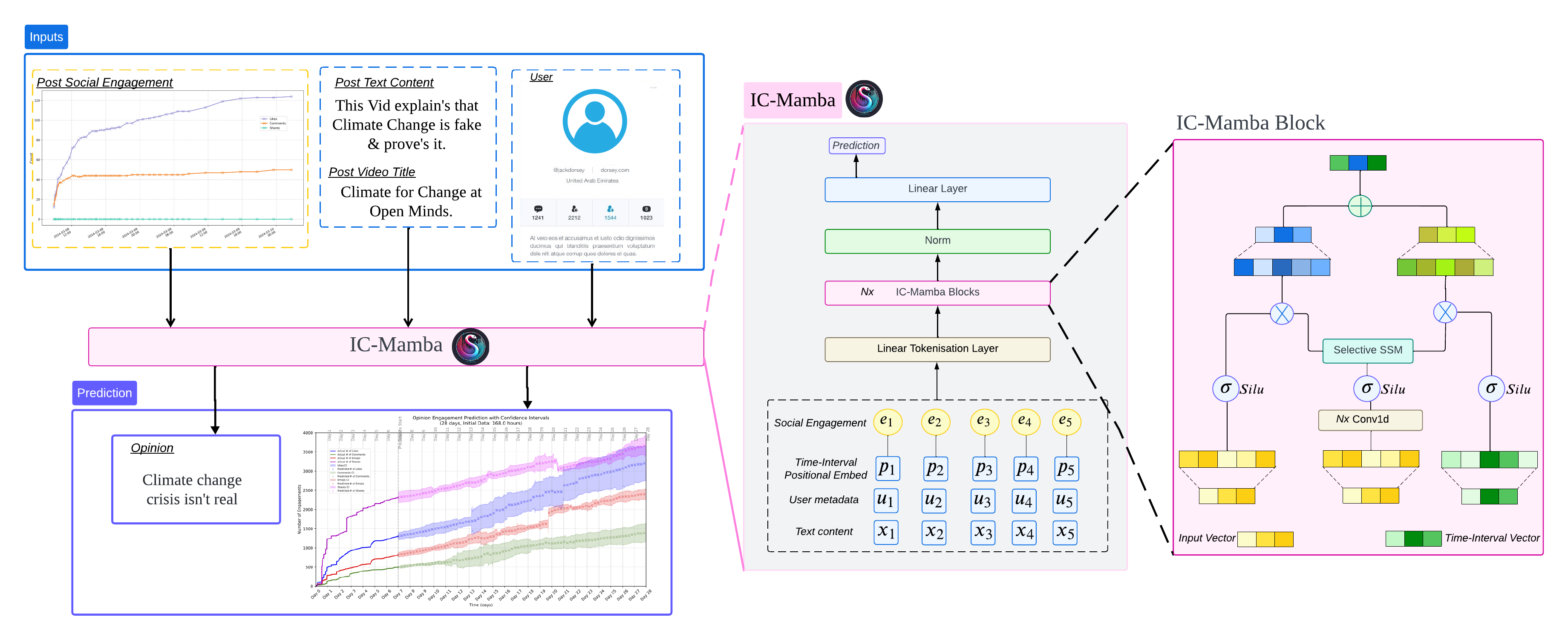}
  \caption{
    Overview of the IC-Mamba Architecture for social media engagement prediction. 
    (left panel) The model first takes three types of inputs (interval-censored social engagement, post content, and user metadata). 
    These inputs are tokenized through a linear tokenization layer. 
    The tokenized sequence (combination of temporal embedding, positional embeddings and user embeddings) is processed through N-stacked \icmamba blocks.
    (right panel) Each \icmamba block contains a selective SSM mechanism and parallel Conv1d operations to handle input and time-interval vectors simultaneously. 
    Lastly, the processed features go through normalization and linear layers to generate the final social engagement predictions.
  }
  \label{fig:overall_archi}
\end{figure*}

\section{Interval-Censored Mamba (IC-Mamba)}
\label{sec: ic-mamba-method} 

This section introduces \icmamba, our proposed approach for engagement prediction illustrated in \cref{fig:overall_archi}.
We begin with the problem statement (\cref{subsec:problem-statement}) and then detail the key components of our architecture:
the time-aware positional embeddings (\cref{subsec:time-aware-embed}), 
the content and sequence embeddings (\cref{subsec:context-sequence-embed}), 
interval-censored state space modeling (\cref{subsec:interval-censored-SSM}), 
the pretraining strategies (\cref{subsec:ic-mamba-pretrain}), and 
the two-tier architecture that enables predictions at both post and opinion levels (\cref{subsec:two-tier-arch}).

\subsection{Problem Statement}
\label{subsec:problem-statement}
Let $\mathcal{E}$ denote a social outbreak event with associated posts $\mathcal{P} = \{p_1, p_2, \dots, p_N\}$. For each post $p \in \mathcal{P}$, we define a tuple $(t_0, x, u, o, H)$ where $t_0$ denotes the original posting time; $x$ represents the textual content; $u$ captures the user metadata; $o \in \mathcal{O}$ indicates the opinion class from the set of possible opinions $\mathcal{O}$; and the interval-censored engagement history is defined as $H = \{(t_j, e_j)\}_{j=1}^{m}$, with 
$m$ as the total number of observation intervals. Each $e_j$ is a $d$-dimensional vector capturing different types of engagement at observation time $t_j$, with intervals $\Delta t_j = t_{j+1} - t_j$ between consecutive observations -- see also \cref{fig:sample_ic_mamba} for how these quantities interact.
See \cref{tab:notations} for a complete reference of mathematical notations used in this work.

Given an observation window $\tau_{obs}$ (e.g., 1 day), let $H_{\tau_{obs}}(p) = \{(t, e) \in H \mid t_0 \leq t \leq t_0 + \tau_{obs}\}$ denote the initial interval-censored engagement history. Let $\Delta t$ be a fixed time interval (e.g.,\ 5 minutes) and $T$ be the prediction horizon (e.g.,\ 28 days). Our goal is to predict the engagement trajectory at regular intervals: $\{\hat{e}(t_0 + \tau_{obs} + k\Delta t)\}_{k=1}^{K}$, where $K = \lfloor T/\Delta t \rfloor$ represents the number of prediction points.


Using this setup, we address two primary tasks. 
(1) Social Engagement Prediction: We predict engagement at both individual and collective levels. 
\emph{Post level}: Predict the engagement trajectory $\hat{e}(t_0 + \tau_{\text{obs}} + k\,\tau_{\text{step}})_{k=1}^{K}$ at regular intervals $\tau_{\text{step}}$ up to horizon $T$ (with $K = \lfloor T/\tau_{\text{step}} \rfloor$), as well as the total cumulative engagement over $T$. 
\emph{Opinion level:} For a given opinion \(o\), predict the collective trajectory ${\hat{E}_o(t_0 + \tau_{obs} + k\tau_{\text{step}})}_{k=1}^{K}$ , where $\hat{E}_o$ is the sum of engagements across all posts $\mathcal{P}_o$ expressing $o$. 
(2) Opinion Classification: We learn a mapping $f: (x, u, H_{\tau_{obs}}) \mapsto \mathcal{O}$ that assigns a post to an opinion class based on its content $x$, user metadata $u$, and engagement history $H_{\tau_{obs}}$.

\subsection{Time-aware Positional Embeddings}
\label{subsec:time-aware-embed}

The temporal dynamics of social media engagement operate at multiple scales -- from rapid initial spread to long-term influence patterns. 
To capture these multi-scale dynamics, we introduce a dual strategy featuring Relative Temporal Encoding (RTE) and Absolute Temporal Encoding (ATE).
RTE captures temporal relationships between two time points $t$ and $t_{ref}$ as
$RTE(t, t_{ref}) = \sin\left(\frac{t - t_{ref}}{\sigma}\right)$, where $\sigma$ is a learnable parameter that allows the model to adapt to varying engagement velocities.
ATE is capturing predictions to the global event timeline by mapping each time point $t$ into a sinusoidal embedding space:
\begin{equation*}
ATE(t) = \left[\sin\left(\frac{t}{10000^{2i/d}}\right), \cos\left(\frac{t}{10000^{2i/d}}\right)\right]_{i=0}^{d/2-1}.
\end{equation*}

These embeddings combine through a learnable projection:
\begin{equation*}
PE(t, t_{ref}) = W_p \begin{bmatrix} RTE(t, t_{ref}) \\ ATE(t) \end{bmatrix},
\end{equation*}
which is then modulated by observed engagement
$EPE(t, t_{ref}, e) = PE(t, t_{ref}) \odot \bigl(1 + \log\left(1 + e\right)\bigl)$,
 where $\odot$ denotes element-wise multiplication, and $e$ is the engagement vector at time $t$.

This engagement-sensitive embedding enables the model to learn characteristic temporal patterns associated with different levels of social impact.
For each post $p \in \mathcal{P}$ and a prediction time $\tau_k$, we construct a \textit{time-aware embedding sequence} $TE^k(p) \in \mathbb{R}^{(m_k + 1) \times d}$ as $
TE^k(p) = \left[EPE(t_{j}, \tau_k, e_{j}) \mid (t_j, e_j) \in H_{\tau_{obs}}(p)\right] \cup \left[PE(\tau_k, \tau_k, 0)\right]$, where $H_{\tau_{obs}}(p) = \{(t, e) \in H \mid t_0 \leq t \leq t_0 + \tau_{obs}\}$ is the observed engagement history within the observation window $\tau_{obs}$.

\subsection{Content and Sequence Embedding}
\label{subsec:context-sequence-embed}

To create a unified representation of social media posts, we must handle both textual content and temporal patterns. 
We use a byte-level BPE tokenizer~\citep{black2022gptneoxb} to process the social media text, enabling us to embed the multi-modal information (content, user metadata, and temporal dynamics) into a single sequence representation:
$SE(p) = {Encoder}([CLS] \oplus [x] \oplus [SEP] \oplus [u] \oplus [SEP] \oplus [T] \oplus [SEP] \oplus [{e_j}])$.
Here, $Encoder$ is a transformer-based function, $x$ is the post text, $u$ is user metadata, $T = \{t_0, t_1, \dots, t_{m}\}$ is the post's timeline of engagement events, $\{e_j\}$ are engagement counts, and $[CLS]$ and $[SEP]$ are special tokens.
Note that the $Encoder$ function maps the input sequence to a fixed-dimensional space $\mathbb{R}^d$, where $d$ is the embedding dimension. 
This allows for building uniform representations regardless of the posts' content or engagement history length.

\subsection{Interval-Censored State Space Modeling}
\label{subsec:interval-censored-SSM}

Here, we extend the Mamba architecture to incorporate time intervals within the state space model. 
Standard SSMs assume regular sampling intervals, which fails to capture social media engagement's irregular and censored nature (see \cref{fig:sample_ic_mamba}). 
We address this through three key components: interval-aware state representation, time-dependent transitions, and selective state updates.

\noindent\textbf{Interval-aware State Representation.}
For each observation time $t_j$ in the engagement history $H_{\tau_{obs}}(p)$, we construct an interval-aware vector $v_j \in \mathbb{R}^{4d}$:
\begin{equation*}
        v_j = [\Delta t_j^-; \log(1 + e_j); \Delta t_j^+; \log(1 + \hat{e}_{j+1})],
\end{equation*}
where $\Delta t_{j}^- = t_{j} - t_{j-1}$ captures the time since the last observation, 
$e_{j}$ is the current engagement vector, 
$\Delta t_{j}^+ = t_{j+1} - t_{j}$ is the forward interval length, and 
$\hat{e}_{j+1}$ is the predicted next engagement vector.

To maintain a consistent representation when transitioning from variable-length historical intervals to fixed-length prediction intervals, at each prediction time point $\tau_k$, we construct: $v_k = [\tau_k - t_j; \log(1 + e_j); \tau_{k+1} - \tau_k; \log(1 + \hat{e}_k)]$,
using the last observed engagement $(t_{j}, e_{j})$ in $H_{\tau_{obs}}$.

\noindent\textbf{Time-Dependent State Transitions.}
We handle varying-length censored intervals by modifying the standard SSM architecture to incorporate time-dependent state transitions. 
For a hidden state dimension $D_h$ and input dimension $D$, our model becomes:
\begin{align*}
    \mathbf{A}_t(\Delta t) &= \exp(\Delta t \cdot \tilde{\mathbf{A}}_t) \in \mathbb{R}^{D_h \times D_h}, \\ 
    \mathbf{h}_t &= \mathbf{A}_t(\Delta t) \mathbf{h}_{t-1} + \mathbf{B}_t \mathbf{x}_t, \quad\quad \mathbf{y}_t = \mathbf{C}_t^T \mathbf{h}_t,
\end{align*}
where $\mathbf{h}_t \in \mathbb{R}^{D_h}$ is the hidden state at time $t$, $\mathbf{x}_t \in \mathbb{R}^D$ is derived from the interval-aware vector $v_j$, and the matrix exponential $\exp(\Delta t \cdot \tilde{\mathbf{A}}_t)$ enables smooth interpolation across censored intervals.

\noindent\textbf{Selective State Processing.}
We integrate the temporal embeddings (${TE}^k(p)$) and interval-aware vectors through parallel pathways:
\begin{equation*}
        [\mathbf{X}, \boldsymbol{\Delta}, \mathbf{B}, \mathbf{C}] = \text{Projection}\left(\mathbf{V}, {TE}^k(p)\right) \in \mathbb{R}^{L \times (D + 1 + 2N)}
\end{equation*}
where $L$ is the sequence length, $\mathbf{V} \in \mathbb{R}^{L \times 4d}$ is the sequence of interval-aware vectors, and ${TE}^k(p)$ provides temporal context. 
The selective SSM mechanism then processes as follows:
\begin{equation*}
\mathbf{Y} = \text{SSM}(\tilde{\mathbf{A}}, \mathbf{B}, \mathbf{C}, \mathbf{X}, \boldsymbol{\Delta}) \in \mathbb{R}^{L \times D},
\end{equation*}

The final output is modulated through a gating mechanism:
\begin{equation*}
\text{Output} = \mathbf{Y} \odot \sigma \bigl(\text{Conv1d}(\mathbf{X})\bigl) \in \mathbb{R}^{L \times D},
\end{equation*}
where $\sigma$ is the Silu activation function~\citep{elfwing2018sigmoid} and Conv1d~\citep{gu2021combining} captures local engagement patterns.

\subsection{IC-Mamba Pretraining}
\label{subsec:ic-mamba-pretrain}

Creating labeled sets of misinformation and disinformation campaigns is a human-time-intensive process, and often, the resulting training sets are too small to allow training an architecture such as \icmamba from scratch.
\cref{alg:pretraining} outlines the pretraining procedure for \icmamba.
We introduce $\mathcal{D} = \{(p_i, H_i, x_i, u_i)\}_{i=1}^M$, a pretraining dataset comprising $1.78$ million posts and their associated social engagement timelines -- totaling over $153$ million timelines -- collected from the two datasets SocialSense~\citep{kong2022slipping} and DiN (detailed in Section~\ref{subsec:datasets}).
Here $M$ is the number of posts and $H_i = \{(t_{i,n}, e_{i,n})\}_{n=1}^{m_i}$ with $|H_i| = m_i$ represents the complete engagement history for post $p_i$.


\noindent\textbf{Objective Function.}
We define two objective functions that we combine for pretraining.

\emph{Engagement Prediction Loss.} 
For each post, we train the model to predict the next engagement vector:
\begin{equation}
  \label{eq:loss_next_pred}
        \mathcal{L}_\text{pred} = \frac{1}{|\mathcal{P}|} \sum_{p \in \mathcal{P}} \sum_{j=0}^{m-1} \|\hat{e}_{j+1} - e_{j+1}\|^2 \enspace,
\end{equation}
where $\hat{e}_{j+1} \in \mathbb{R}^d$ is the predicted engagement vector.

\emph{Temporal Coherence Loss.} 
We enforce consistent state transitions across intervals:
\begin{equation}
  \label{eq:loss_interval}
        \mathcal{L}_\text{temp} = \frac{1}{|\mathcal{P}|} \sum_{p \in \mathcal{P}} \sum_{j=0}^{m-1} \|\mathbf{h}_{j+1} - \exp(\Delta t_j^+ \cdot \tilde{\mathbf{A}}_t)\mathbf{h}_j\|^2 \enspace,
\end{equation}
where $\mathbf{h}_j \in \mathbb{R}^{D_h}$ is the hidden state at time $t_j$ and the exponential term comes from our SSM formulation.

The pretraining loss combines these objectives from \cref{eq:loss_next_pred} and \cref{eq:loss_interval} as $\mathcal{L}_\text{total} = \mathcal{L}_\text{pred} + \lambda \mathcal{L}_\text{temp}$,
where $\lambda$ is a hyperparameter balancing the two losses.
%

\begin{algorithm}[t]
\caption{\icmamba Pretraining}
\label{alg:pretraining}
\begin{algorithmic}[1]
\State Initialize parameters $\theta = \{\tilde{\mathbf{A}}, \mathbf{B}, \mathbf{C}, \mathbf{W}_p, \theta_\text{Encoder}\}$
\For{epoch $= 1$ to $N_\text{epochs}$}
    \For{batch $\mathcal{B} \subset \mathcal{D}$}
        \State Construct interval-aware vectors $\{\mathbf{v}_{j}\}_{j \in \mathcal{B}}$ 
        \State Compute temporal embeddings $\{TE^k(p)\}_{p \in \mathcal{B}}$
        \State $[\mathbf{X}, \boldsymbol{\Delta}, \mathbf{B}, \mathbf{C}] \gets \text{Projection}(\{\mathbf{v}_{j}\}, \{TE^k(p)\})$
        \State $\mathbf{H} \gets \text{SSM}(\tilde{\mathbf{A}}, \mathbf{B}, \mathbf{C}, \mathbf{X}, \boldsymbol{\Delta})$
        \State $\hat{\mathbf{E}} \gets \text{MLP}(\mathbf{H})$
        \State Compute $\mathcal{L}_\text{pred}$ (\cref{eq:loss_next_pred}) and $\mathcal{L}_\text{temp}$ (\cref{eq:loss_interval})
        \State Update $\theta$ using $\nabla_\theta(\mathcal{L}_\text{pred} + \lambda \mathcal{L}_\text{temp})$
    \EndFor
\EndFor
\State \Return $\theta$
\end{algorithmic}
\end{algorithm}

\subsection{Two-Tier \icmamba Architecture}
\label{subsec:two-tier-arch}

\begin{figure}[t]
  \centering
  \includegraphics[width=0.85\linewidth]{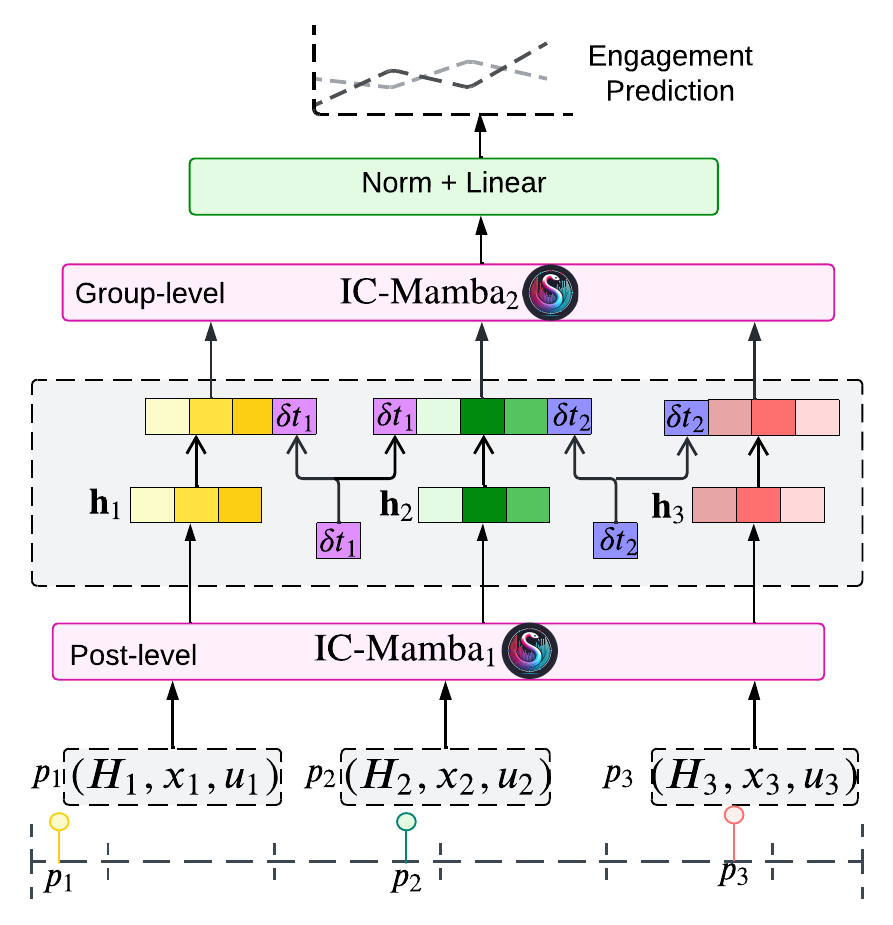}
  \caption{
    Two-Tier \icmamba Architecture. The bottom-tier model ($\text{IC-Mamba}_{1}$) learns post-level representations from historical ($H$), content ($x$), and user ($u$) features, while the top-tier model ($\text{IC-Mamba}_2$) captures temporal dependencies across intervals $\delta t$ to jointly predict individual post virality and aggregate narrative engagement dynamics.
  }
  \label{fig:2tier_ic-mamba}
\end{figure}

It is desirable to model and predict the engagement dynamics of a group of posts expressing the same opinion -- dubbed \emph{the engagement of an opinion}.
We propose a hierarchical two-tier architecture, showcased in \cref{fig:2tier_ic-mamba}.
The intuition of the two-tier \icmamba model is that the first tier ($\text{IC-Mamba}_{1}$) models the arrival of engagement on an individual post. 
The second tier ($\text{IC-Mamba}_2$) models the arrival of posts within an opinion.

\noindent\textbf{Post-Level Processing.}
In the first tier, for each opinion $o$, we process all posts $p_i \in \mathcal{P}_o$ individually using the $\text{IC-Mamba}{1}$ model:
\begin{equation}
\label{eq:tier-1-icmamba}
        \mathbf{h}_i = \text{IC-Mamba}_{1}(H_{\tau_{obs}}(p_i), x_i, u_i), \quad \forall p_i \in \mathcal{P}_o
\end{equation}
where $H{\tau_{obs}}(p_i)$ is the interval-censored engagement over observation window and $\mathbf{h}_i$ the hidden state representation of post $p_i$.

\noindent\textbf{Group-Level Dynamics.}
In the second tier, we model the temporal interactions between posts sharing opinion $o$. 
By ordering posts in $\mathcal{P}_o$ chronologically by posting time $t_i^{\mathrm{p}}$, we capture the inter-post intervals $\delta t_i = t_{i+1}^{\mathrm{p}} - t_i^{\mathrm{p}}$ between posts in the group. 
The group-level dynamics are modeled using $\text{IC-Mamba}_2$ with $\mathbf{h}_i$ from \cref{eq:tier-1-icmamba}:
\begin{equation*}
\mathbf{z}_o = \text{IC-Mamba}_2({(\mathbf{h}_i, \delta t_i)}).
\end{equation*}



\begin{figure*}[t]
    \centering
    \newcommand\myheight{0.138}
    \subfloat[]{
        \includegraphics[height=\myheight\textheight]{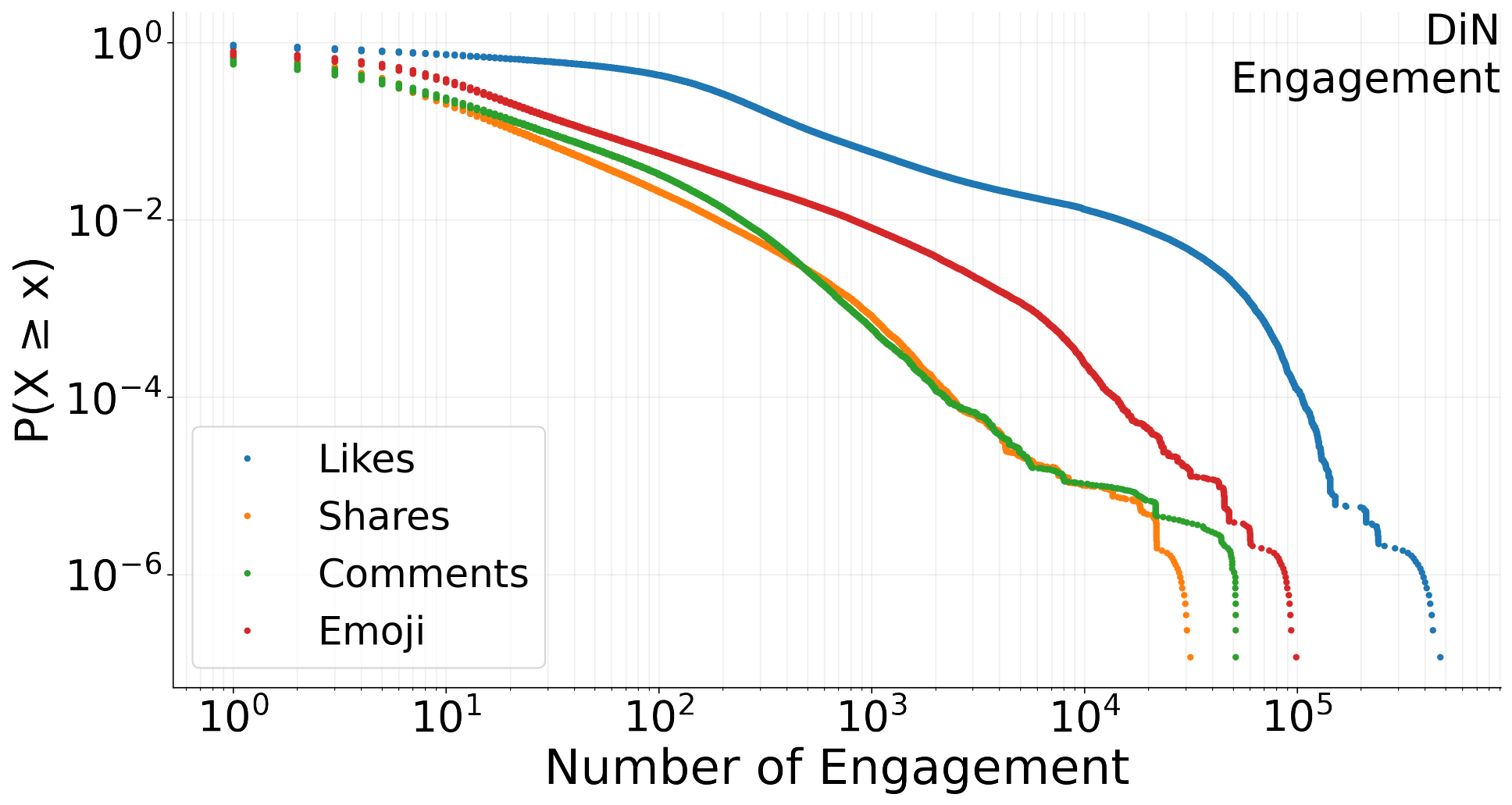}
        \label{fig:data_insights_DiN}
    }%
    \subfloat[]{
        \includegraphics[height=\myheight\textheight]{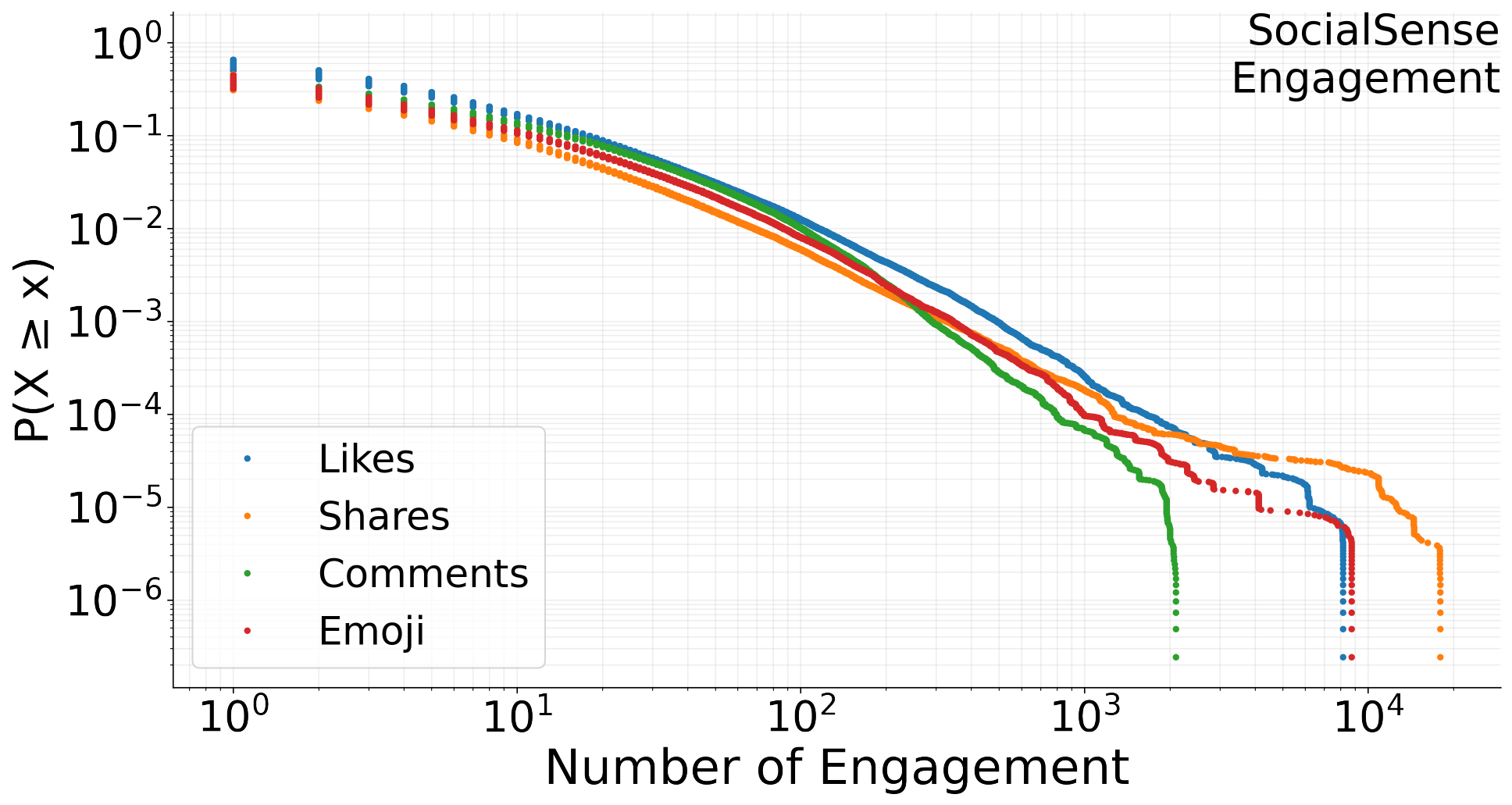}
        \label{fig:data_insights_cc}
    }%
    \subfloat[]{
        \includegraphics[height=\myheight\textheight]{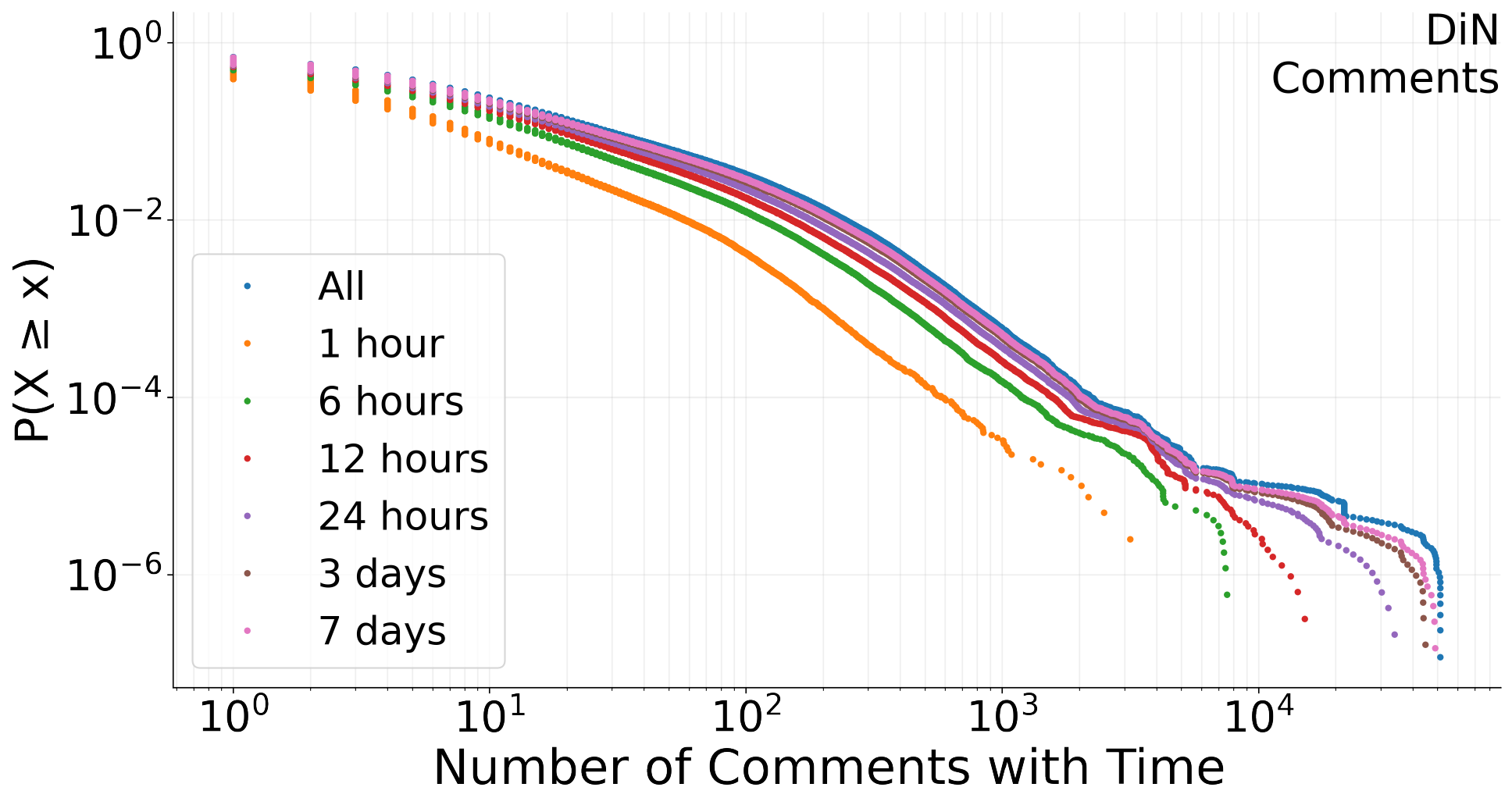}
        \label{fig:data_insights_comments}
    }%
    \caption{ Engagement distribution patterns across social media content. (a) Log-scale ECCDF of engagement metrics for the DiN dataset. (b) Log-scale ECCDF of engagement metrics from the climate change theme in SocialSense. (c) Temporal evolution of comment distributions across different time windows ranging from 1 hour to 7 days. Note: ECCDF represents Empirical Complementary Cumulative Distribution Functions.}
    \label{fig:data_insights}
\end{figure*}
\section{Experiments and Results}
In this section, we present the experimental setup and the results we obtain; 
including datasets and data insights (\cref{subsec:datasets}), 
the baseline models we compare against (\cref{subsec:baselines-xp-setup}), and 
the results that address our research questions (\cref{subsec:results}).
\begin{table}[t]
    \caption{Dataset Statistics}
    \setlength{\tabcolsep}{3pt}
    \centering
    \begin{tabular}{@{}lrrrr|r@{}}
        \toprule
        \textbf{Dataset} & \multicolumn{1}{c}{Bushfire} & \multicolumn{1}{c}{Climate Ch.} & \multicolumn{1}{c}{Vaccin.} & \multicolumn{1}{c|}{COVID-19} & \multicolumn{1}{c}{DiN} \\ \midrule
        \textbf{\#posts}                     & 78,030                       & 138,278                            & 178,894                         & 640,100                       & 746,653                 \\
        \textbf{\#users}                     & 13,438                       & 25,850                             & 34,652                          & 67,727                        & 41                      \\
        \textbf{\#opinions}                  & 15                           & 24                                 & 27                              & 17                            & 9                       \\ \bottomrule
    \end{tabular}
    \label{tab:data_statistics}
\end{table}
\subsection{Datasets}
\label{subsec:datasets}


\paragraph{Datasets}
Our experiments use two Facebook datasets: the theme-focused SocialSense dataset~\citep{kong2022slipping} and the user-centric Disinformation Network (DiN) dataset.
For each post in our datasets, we collect historical engagement metrics (likes, shares, comments, emoji reactions) collected via CrowdTangle API\footnote{\url{https://www.crowdtangle.com/} before its termination in August 2024.}. 
\emph{SocialSense} contains posts and comments from four main themes during 2019-2021(see \cref{tab:data_statistics}) that attracted significant volumes of misinformation and conspiratorial discussions. 
The \emph{DiN dataset} comprises posts from $41$ accounts (2019-2024). Social science experts systematically analyzed and assigned narrative labels to these posts through comprehensive content evaluation to detect suspected coordinated information operations. 
The two datasets capture the dynamics of misinformation across diverse real-world events (SocialSense) and disinformation narrative spread by information operation networks (DiN).~\footnote{Note that, posts with fewer than four engagement intervals were excluded from model evaluation to ensure sufficient temporal depth. }

\paragraph{Data Insights}
\cref{fig:data_insights}(a) and (b) present the Empirical Complementary Cumulative Distribution Functions (ECCDFs) for likes, shares, comments, and emoji reactions across DiN (a) and the Climate Change theme in SocialSense (b). The survival probability $P(X \geq k)$ measures the likelihood of achieving at least $k$ engagements \citep{clauset2009power}, and the power-law exponent $\alpha$ characterizes the decay rate \citep{newman2005power}. While Climate Change content rarely exceeds $10^4$ total engagements, DiN reaches $10^6$, indicating significantly broader reach.
In the low-engagement regime ($1 \leq k \leq 10$), DiN exhibits a higher survival probability ($\alpha \approx 2.1$) compared to Climate Change ($\alpha \approx 2.4$), suggesting stronger early visibility potential. The mid-range ($10 \leq k \leq 1000$) shows uniform decay across engagement types for Climate Change, reflecting organic interaction patterns. In contrast, DiN reveals marked stratification, especially in likes. Beyond $k > 1000$, Climate Change content plateaus near $10^3$ engagements, aligning with established social network theory regarding human-scale constraints -- approximately 150 stable connections, known as Dunbar’s number \citep{dunbar1992neocortex} -- while DiN content transcends these natural limits, reaching $10^6$ engagements.

\cref{fig:data_insights}(c) offers examines comment distributions over time windows ranging from one hour to seven days. The scale-invariant, power-law structure persists across all observation periods, though longer windows ($3$–$7$ days) exhibit slightly elevated survival probabilities beyond $10^3$. This self-similar temporal behavior distinguishes naturally diffusing, high-visibility content from artificially amplified patterns, underscoring the unique viral longevity of DiN.

\begin{table*}[tb]
    \caption{
        Post-level engagement prediction performance of \icmamba vs baselines on SocialSense (four themes) and DiN; 
        measured using RMSE and MAPE (lower is better), and $R^2$ (higher is better).
        Best performance in boldface.
    }
    \centering
        \begin{adjustbox}{max width=1.0\linewidth}
            \begin{tabular}{lc@{\;\;}c@{\;\;}cc@{\;\;}c@{\;\;}cc@{\;\;}c@{\;\;}cc@{\;\;}c@{\;\;}cc@{\;\;}c@{\;\;}c}
                \toprule
                \toprule
                \multirow{2}{*}{Model}                    & \multicolumn{3}{c}{Bushfire} & \multicolumn{3}{c}{Climate Change} & \multicolumn{3}{c}{Vaccination} & \multicolumn{3}{c}{COVID-19} & \multicolumn{3}{c}{DiN}                                                                                                                                                                           \\
                \cmidrule(l{.75em}r{.75em}){2-4}\cmidrule(l{.75em}r{.75em}){5-7}\cmidrule(l{.75em}r{.75em}){8-10}\cmidrule(l{.75em}r{.75em}){11-13} \cmidrule(l{.75em}r{.75em}){14-16}
                                                          & RMSE                         & MAPE                               & $R^2$                           & RMSE                      & MAPE                    & $R^2$          & RMSE           & MAPE           & $R^2$          & RMSE           & MAPE           & $R^2$          & RMSE           & MAPE           & $R^2$          \\ \toprule
                TSTransformer~\cite{vaswani2017attention} & 0.185                        & 0.232                              & 0.651                           & 0.192                     & 0.241                   & 0.643          & 0.180          & 0.226          & 0.658          & 0.188          & 0.236          & 0.647          & 0.221          & 0.276          & 0.568          \\
                Informer~\cite{zhou2021informer}          & 0.172                        & 0.216                              & 0.678                           & 0.179                     & 0.224                   & 0.670          & 0.167          & 0.210          & 0.685          & 0.175          & 0.220          & 0.674          & 0.206          & 0.258          & 0.598          \\
                Autoformer~\cite{wu2021autoformer}        & 0.163                        & 0.204                              & 0.697                           & 0.169                     & 0.212                   & 0.689          & 0.158          & 0.198          & 0.704          & 0.166          & 0.208          & 0.693          & 0.195          & 0.244          & 0.619          \\

                \midrule
                MBPP~\cite{rizoiu2022interval}             & 0.183                        & 0.229                              & 0.655                           & 0.192                     & 0.241                   & 0.643          & 0.181          & 0.227          & 0.656          & 0.189          & 0.237          & 0.645          & 0.222          & 0.278          & 0.566          \\
                IC-TH~\cite{kong2023interval}             & 0.156                        & 0.195                              & 0.712                           & 0.162                     & 0.203                   & 0.704          & 0.151          & 0.189          & 0.719          & 0.159          & 0.199          & 0.708          & 0.187          & 0.234          & 0.636          \\
                \midrule
                TH~\cite{zuo2020transformer}              & 0.149                        & 0.187                              & 0.726                           & 0.155                     & 0.194                   & 0.718          & 0.144          & 0.181          & 0.733          & 0.152          & 0.190          & 0.722          & 0.179          & 0.224          & 0.652          \\
                TS-Mixer~\cite{chen2023tsmixer}           & 0.155                        & 0.194                              & 0.714                           & 0.161                     & 0.202                   & 0.706          & 0.150          & 0.188          & 0.721          & 0.158          & 0.198          & 0.710          & 0.186          & 0.233          & 0.638          \\
                \midrule
                Mamba~\cite{mamba2}          & 0.124                        & 0.155                              & 0.776                           & 0.129                     & 0.161                   & 0.768          & 0.119          & 0.149          & 0.783          & 0.127          & 0.159          & 0.772          & 0.150          & 0.188          & 0.708          \\
                \icmamba w/o text                         & 0.123                        & 0.154                              & 0.778                           & 0.128                     & 0.160                   & 0.770          & 0.118          & 0.148          & 0.785          & 0.126          & 0.158          & 0.774          & 0.145          & 0.181          & 0.718          \\
                \icmamba w/o user                         & 0.120                        & 0.150                              & 0.785                           & 0.125                     & 0.156                   & 0.777          & 0.115          & 0.144          & 0.792          & 0.123          & 0.154          & 0.781          & 0.146          & 0.183          & 0.716          \\
                \icmamba w/o time                         & 0.121                        & 0.151                              & 0.783                           & 0.126                     & 0.157                   & 0.775          & 0.116          & 0.145          & 0.790          & 0.124          & 0.155          & 0.779          & 0.149          & 0.186          & 0.711          \\

                \icmamba                                  & \textbf{0.118}               & \textbf{0.148}                     & \textbf{0.789}                  & \textbf{0.123}            & \textbf{0.154}          & \textbf{0.781} & \textbf{0.113} & \textbf{0.142} & \textbf{0.796} & \textbf{0.121} & \textbf{0.152} & \textbf{0.785} & \textbf{0.143} & \textbf{0.179} & \textbf{0.723} \\
                \bottomrule
                \bottomrule
            \end{tabular}
        \end{adjustbox}
    \label{tab:postlevel_results}
\end{table*}

\subsection{Baselines and Experimental Setup}
\label{subsec:baselines-xp-setup}

We compare our \icmamba model against the following state-of-the-art baselines, including generative models, transformer-based architectures and state space models:
\begin{itemize}[leftmargin=*]
    \setlength\itemsep{0em}
    \item \textbf{TimeSeriesTransformer}~\cite{li2019enhancing}\footnote{\url{https://huggingface.co/docs/transformers/en/model_doc/time_series_transformer}} is a transformer-based model specifically adapted for time series forecasting. It applies self-attention mechanisms to capture temporal dependencies.
    \item \textbf{Informer}~\cite{zhou2021informer}\footnote{\url{https://huggingface.co/docs/transformers/en/model_doc/informer}} is a long sequence time-series forecasting model that uses a ProbSparse self-attention mechanism to handle long-term dependencies.
    \item \textbf{Autoformer}~\cite{wu2021autoformer}\footnote{\url{https://huggingface.co/docs/transformers/en/model_doc/autoformer}} is a decomposition-based architecture for long-term time series forecasting. It uses an auto-correlation mechanism to identify period-based dependencies and a series decomposition architecture for trend-seasonal decomposition.
    \item \textbf{Mean Behaviour Poisson (MBP)}~\cite{rizoiu2022interval} is a generative time series model that uses a compensator function to model non-linear engagement patterns. It treats each engagement as an event in a continuous time process and optimizes post-specific parameters to model the expected cumulative engagements over time to capture the growth patterns.
    \item \textbf{Transformer-Hawkes (TH)}~\cite{zuo2020transformer} is a model that combines the transformer architecture with the Hawkes process for modeling sequential events. It uses self-attention mechanisms to capture temporal dependencies in event sequences.
    \item \textbf{Interval-Censored Transformer Hawkes (IC-TH)}~\cite{kong2023interval} is a TH extension designed to handle interval-censored data. It adapts the transformer architecture to work with event data where exact occurrence times are unknown but bounded within intervals.
    \item \textbf{TS-Mixer}~\cite{chen2023tsmixer}\footnote{\url{https://github.com/google-research/google-research/tree/master/tsmixer}} is a model that combines MLPs and transformers for time series forecasting. It uses separate mixing operations across the temporal and feature dimensions, allowing it to capture both temporal patterns and feature interactions.
    \item \textbf{Mamba}~\cite{mamba2}\footnote{\url{https://huggingface.co/docs/transformers/en/model_doc/mamba2}} is a selective state space model, it uses selective algorithms instead of attention mechanisms for sequence modeling. It can handle long-range dependencies in sequential data for time series analysis tasks.
\end{itemize}
\noindent\textbf{Experimental settings.}
We use a temporal holdout evaluation protocol across all the datasets.
We chronologically order all posts and use the earliest $70\%$ for training, the next $15\%$ for validation, and the most recent $15\%$ for testing. 
This ensures no future information leaks into training and models are evaluated on their ability to generalize to future posts.
Models are implemented using PyTorch, with hyperparameters and other settings detailed in \cref{sec:exp_settings}.

\subsection{Engagement Prediction--RQ1}
\label{subsec:results}

We evaluate the performance of our models with two tasks: \emph{engagement forecasting} and \emph{opinion classification}. 
For \emph{engagement prediction}, we observe the first six hours of engagement metrics for each post and forecast the overall engagement metrics (i.e.,\ at $T = \infty$). 
For \emph{opinion classification}, we evaluate our model's classification performance at multiple granularities.
We perform a \emph{post-level opinion classification} across the four SocialSense themes (\textit{bushfire}, \textit{climate change}, \textit{vaccination}, and \textit{COVID-19}) -- that is, we predict if a given post expresses one of the predefined opinions.
For the DiN dataset, we perform a \emph{user-level opinion classification} --
classify the presence of opinions across multiple posts from the same user.

\noindent\textbf{Post-level Engagement Prediction Performance -- RQ1}
\cref{tab:postlevel_results} reports the performance metrics using three standard measures.
We evaluate the models using RMSE to assess absolute prediction errors (crucial for high-engagement posts), MAPE for scale-independent accuracy, and $R^2$ to measure explained variance in engagement predictions.
\icmamba outperforms all baselines on every metric (RMSE, MAPE, and $R^2$) and dataset, while the original Mamba architecture ranks consistently second, confirming the effectiveness of state space models. Among transformers, IC-TH improves upon TH, and TS-Mixer outperforms both Autoformer and Informer; TSTransformer lags behind. Interestingly, the lightweight MBP model, still competes well on some events (particularly \textit{bushfire} and \textit{climate change}).
All models exhibit performance degradation on the DiN dataset, reflecting the complexity of predicting engagement in coordinated campaigns. For models supporting dynamic prediction time points, additional results for next-time and next social engagement metrics are provided in Appendix~\ref{sec:next_token_pred}.

We conduct an ablation study to understand the contribution of different components by removing text, user, and temporal features from \icmamba (\cref{tab:postlevel_results}). 
Text features demonstrate a stronger influence on SocialSense datasets, where their removal leads to a 0.005 RMSE increase, compared to a smaller 0.002 RMSE increase in the DiN dataset. This difference highlights the crucial role of textual content in organic content spread versus coordinated campaigns. Temporal features, conversely, show greater impact on the DiN dataset, where their removal results in a 0.006 RMSE increase, compared to a 0.003 RMSE increase in SocialSense datasets. This may suggest the strategic temporal patterns in coordinated disinformation campaigns. User features maintain consistent importance across both datasets, with their removal causing similar performance degradation (0.002-0.003 RMSE increase) regardless of the dataset type. Even with text features removed, \icmamba still outperforms IC-TH, improving RMSE from 0.156 to 0.123 on the Bushfire dataset, demonstrating the fundamental strength of our model's architectural design.

\noindent\textbf{Opinion-level Classification Performance --RQ1}
In our classification settings, we tackled datasets of varying complexity: the \textit{bushfire} dataset contains $9$ opinions, \textit{climate change} and \textit{vaccination} each have 12 classes, the \textit{COVID-19} dataset includes $10$ classes, and the DiN (Disinformation Narrative) dataset comprises $9$ distinct narrative labels. 
We also include a random classification baseline with an expected F1 score of $1/N$ for each dataset, where $N$ is the number of classes.
Note that we removed opinions with less than $5,000$ posts in this experimental setting.

\cref{tab:cls_results} presents the macro-averaged F1 scores for classification across models and datasets. IC-Mamba consistently outperforms all others, achieving F1 scores between $0.69$ and $0.75$. While BERT performs well on SocialSense (F1: $0.62$–$0.68$), both models see significant drops on DiN, with IC-Mamba scoring $0.52$ and BERT falling to $0.11$. This highlights the limitations of text-only analysis for DiN, where narrative elements demand more complex temporal or contextual understanding.
Informer, Autoformer, and Mamba struggle on SocialSense (F1 < $0.41$) but perform relatively better on DiN, with Mamba achieving its best score of $0.32$. This suggests that temporal and non-textual features are critical for narrative detection, contrasting with the outbreak event focus of SocialSense.

\begin{table}[t]
    \caption{
        Opinion Classification results; F1 scores are reported; higher is better; best results in boldface.}
    \centering
        \begin{adjustbox}{max width=1.0\linewidth}
            \begin{tabular}{l@{\;\;}c@{\;\;}c@{\;\;}c@{\;\;}c@{\;\;}c}
                \toprule
                \toprule
                Model                              & Bushfire       & Climate        & Vaccination    & COVID-19          & DiN            \\ \toprule
                Random                             & 0.111          & 0.083          & 0.083          & 0.083          & 0.111          \\
                \midrule
                Informer~\cite{zhou2021informer}   & 0.323          & 0.291          & 0.274          & 0.299          & 0.248          \\
                Autoformer~\cite{wu2021autoformer} & 0.342          & 0.313          & 0.278          & 0.324          & 0.255          \\
                \midrule
                BERT~\cite{vaswani2017attention}   & 0.676          & 0.652          & 0.621          & 0.644          & 0.107          \\
                \midrule
                Mamba~\cite{mamba2}   & 0.412          & 0.375          & 0.363          & 0.388          & 0.316          \\
                \icmamba                           & \textbf{0.751} & \textbf{0.724} & \textbf{0.687} & \textbf{0.705} & \textbf{0.508} \\
                \bottomrule
                \bottomrule
            \end{tabular}
        \end{adjustbox}
    \label{tab:cls_results}
\end{table}
\begin{figure*}[t]
    \centering
    \newcommand\myheight{0.130}

    \subfloat[]{
        \includegraphics[height=\myheight\textheight]{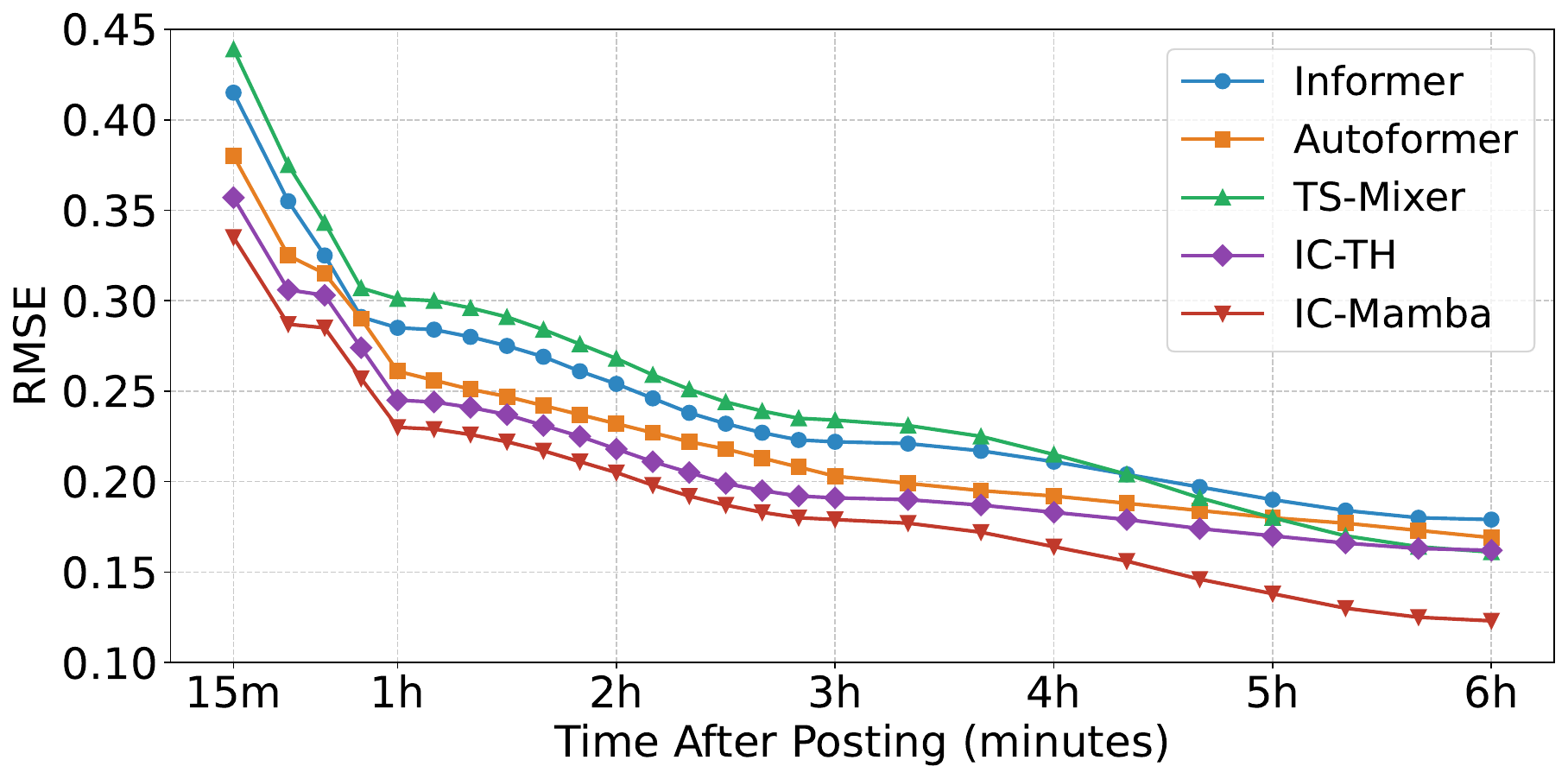}
        \label{fig:early_prediction}
    }%
    \subfloat[]{
        \includegraphics[height=\myheight\textheight]{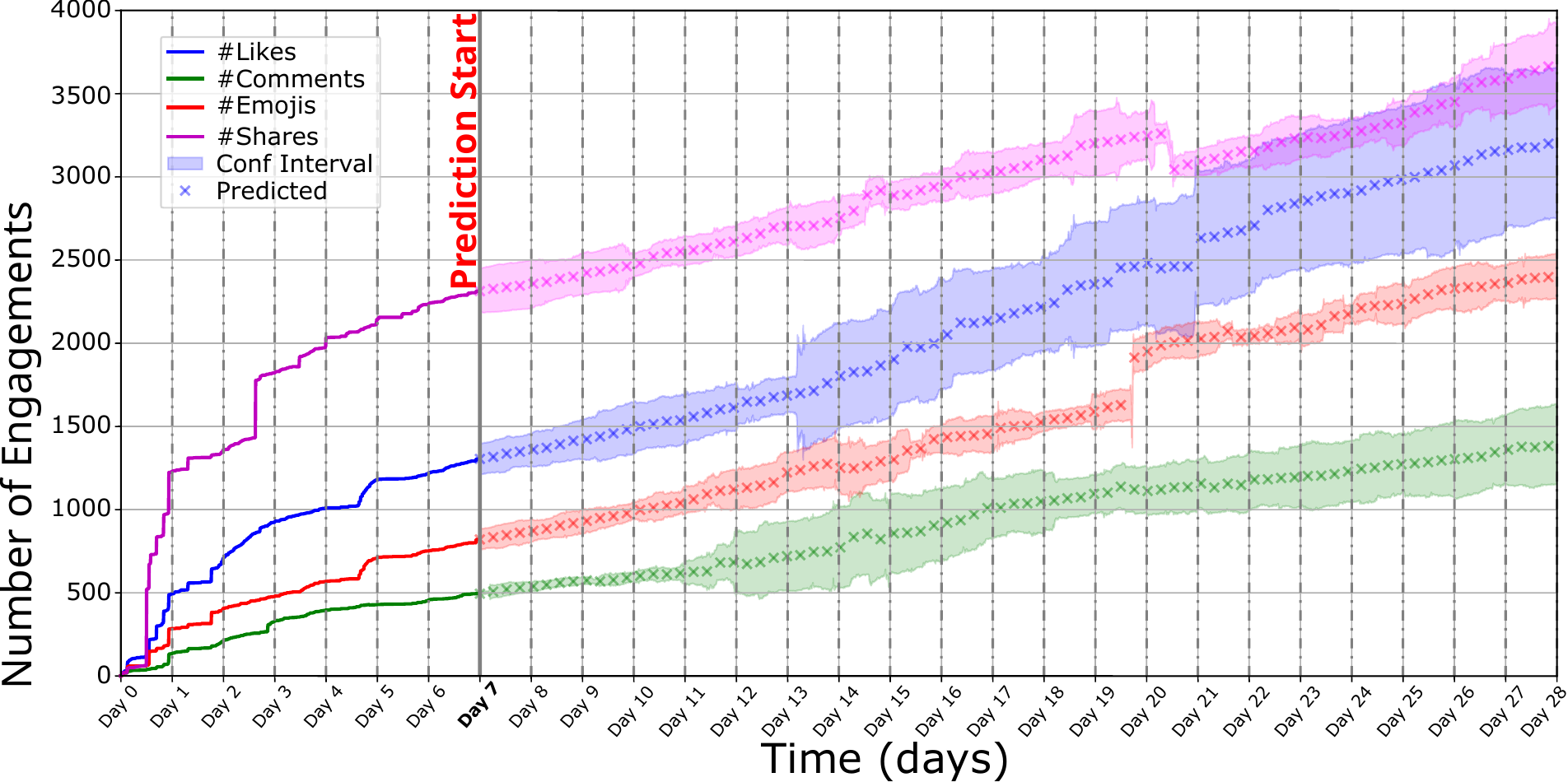}
        \label{fig:dynamic_day7}
    }%
    \subfloat[]{
        \includegraphics[height=\myheight\textheight]{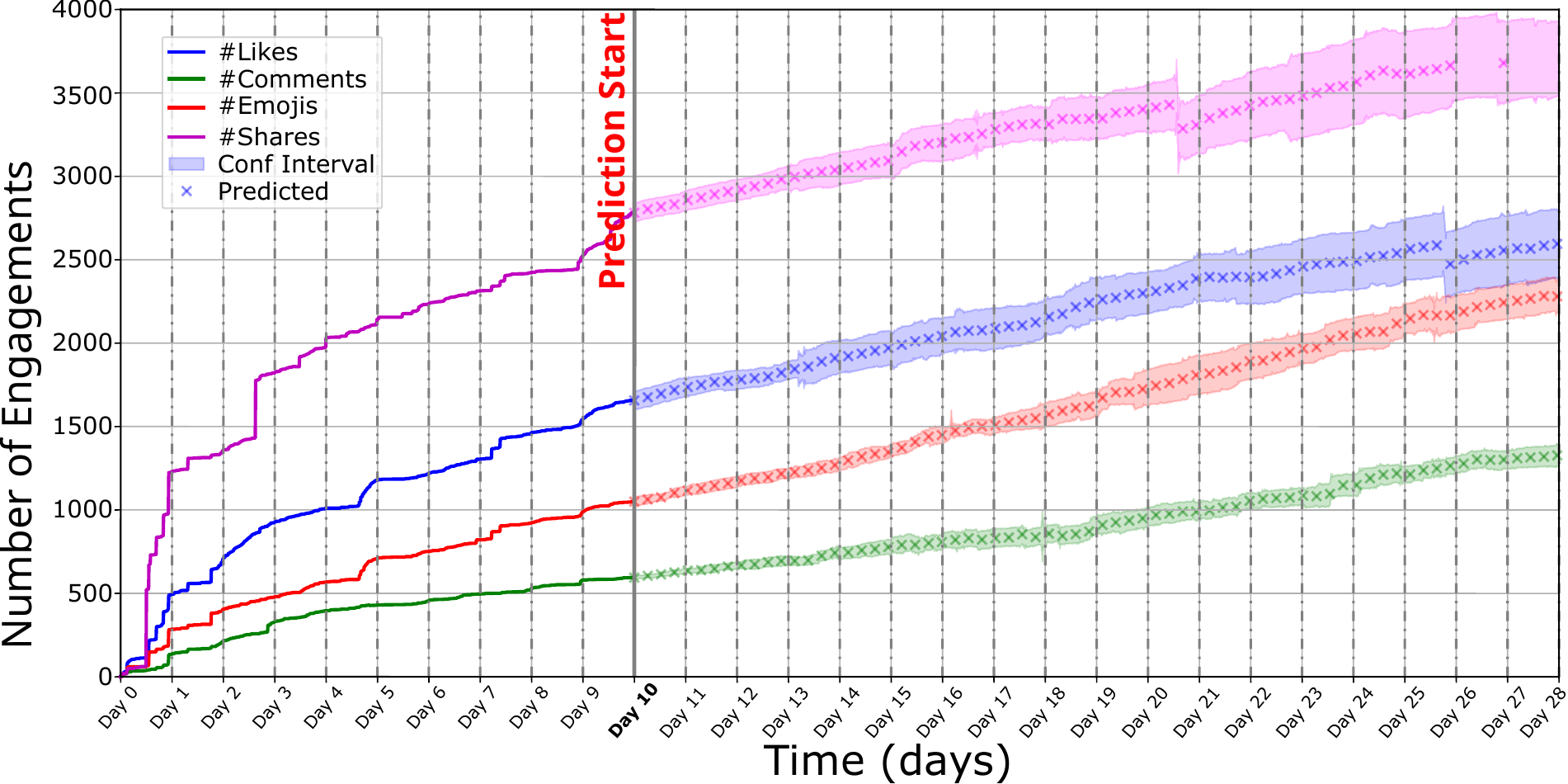}
        \label{fig:dynamic_day10}
    }%
    \caption{
        Comparative analysis of early prediction performance and dynamic forecasting. (a) Performance comparison on RMSE between IC-Mamba and baseline models from 15 minutes to 6 hours after posting. 
        (b)(c) \icmamba's 28-day predictions with 5-minute intervals using 7-day (b) and 10-day (c) input windows respectively.
    }
    \label{fig:dynamic_pred}
\end{figure*}
\subsection{Early Engagement Prediction--RQ2}
\label{sec:early_pred}
We vary the length of the observed period in the temporal holdout setup (see \cref{subsec:baselines-xp-setup})
to assess how well different models can forecast engagement in the critical initial hours after a post is made.
\cref{fig:early_prediction} shows RMSE-based early prediction performance for the climate change theme in SocialSense, measured at intervals from 15 minutes to 6 hours after posting, across Informer, Autoformer, TS-Mixer, IC-TH, and \icmamba.

All models demonstrate substantial improvement in prediction accuracy over time, with error rates decreasing from 15 minutes to 6 hours. The most notable improvements occur in the first hour, particularly between 15-50 minutes, suggesting that the first hour of a post's life is crucial for accurate engagement forecasting.
\icmamba outperforms other models across all time points, and its performance advantage increases over time. While all models show similar patterns of improvement in the first hour, \icmamba continues to achieve increasingly better RMSE scores through the 6-hour mark, reaching the lowest RMSE of 0.118. 
IC-TH maintains second-best performance throughout most of the timeline, followed by Autoformer. The Informer and TS-Mixer models show higher error rates, with their performance plateauing more quickly than the interval-censored approaches. This performance gap may illustrate the benefits of interval-censored modeling in engagement prediction tasks on real-world social media platforms, while the widening gap in RMSE scores over time suggests that \icmamba's improvements go beyond just interval-censored modeling, potentially indicating better long-range dependency learning.

\subsection{Dynamic Opinion-level Prediction--RQ3}
\label{sec:dynamic_pred}

This section simulates a real-world monitoring and forecasting scenario.
We analyze the opinion ``Climate change is a UN hoax'' from the SocialSense climate dataset. 
\cref{fig:dynamic_pred}(b)(c) demonstrates our dynamic prediction approach at opinion-level across multiple interaction types (likes, comments, emojis, and shares) over a 28-day period. We showcase two scenarios of initial data windows -- 168 hours (1 week), and 240 hours (10 days). 
Our model first processes the initial historical data window to establish baseline engagement patterns. As time progresses beyond these initial periods (marked by "Predictions Start" lines), the model continuously incorporates new engagement data to refine its predictions. The shaded areas around each prediction line represent the $95\%$ confidence intervals -- obtained from all previous prediction for this time -- providing a measure of prediction uncertainty over time.
We see that the uncertainty reduces as more initial data is available, suggesting that increased historical data improves the model's predictive accuracy.


\section{Conclusion}

\icmamba demonstrates strong performance in modeling interval-censored engagement data, providing early predictions of viral content, and tracking long-term opinion spread across platforms.
Through the novel integration of interval-censored modeling and temporal embeddings within a state space model, \icmamba achieves strong performance in predicting dynamic misinformation and disinformation engagement patterns and opinion classification.
These capabilities enable platforms and researchers to identify potentially harmful content and coordinated campaigns in their early stages, facilitating proactive intervention strategies while respecting platform constraints and user privacy. Future work could enhance IC-Mamba through cross-platform dynamics modeling, interpretable attention mechanisms, and real-time deployment adaptations.
Online misinformation and information campaigns are part of our digital ecosystem today, but we do not need to resign ourselves to reactively attempting damage control after the fact. With IC-Mamba, we can identify the next QAnon or climate change denialism conspiracies before they gain mass exposure; and we could mitigate the damage it can do to our lives and democratic societies. We provide detailed discussion of ethical considerations and safeguards in \cref{sec:ethics}.

\begin{acks}
  This research was supported by the Advanced Strategic Capabilities Accelerator (ASCA), 
  the Australian Department of Home Affairs, 
  the Defence Science and Technology Group, the Defence Innovation Network.
  the Australian Academy of Science, and 
  the National Science Centre, Poland (Project No. 2021/41/B/HS6/02798).
\end{acks}
\bibliographystyle{ACM-Reference-Format}
\bibliography{ref}

\appendix
\section{Ethics Considerations}
\label{sec:ethics}
Our research exclusively uses publicly available Facebook data on CrowdTangle, adhering to the platform's terms of service and research guidelines. We implement strict data protection measures: all user identifiers are anonymized, personal information is excluded from our analysis, and we focus solely on aggregate engagement patterns. Our data collection and processing procedures have been reviewed and approved. We maintain data minimization principles, collecting only information necessary for our research objectives.

\section{Experimental Settings}
\label{sec:exp_settings}
To evaluate our proposed \icmamba model, we conducted systematic experiments with different parameter configurations. This section includes our experimental environment, hyperparameters, and optimization approach.

\subsection{Experimental Environment}
\label{subsec:exp_env}
All experiments were conducted using PyTorch 2.0 on a GPU cluster with 4xNVIDIA A100 GPUs with 40GB memory.
All reported figures are averaged across ten runs with different random seeds.

\subsection{Hyper-parameters}
\label{subsec:hyper-parameters}
\cref{tab:hyperparameters} lists the hyperparameter ranges used in our experiments.
\begin{table}[htb]
\caption{\icmamba Hyperparameters}
\label{tab:hyperparameters}
\begin{tabular}{llc}
\toprule
\toprule
 & \textbf{Parameter} & \textbf{Range} \\
\midrule
& Hidden Size ($N$) & [256, 1024] \\
& Input Dimension ($D$) & [64, 128] \\
& Number of IC-Mamba Blocks & [2, 8] \\
& Embedding Dimension ($d$) & [128, 512] \\
& Interval-Censor State Dim & [512, 2048] \\
& Sequence Length ($L$) & [1024, 8192] \\
\midrule
& Batch Size & [16, 128] \\
& Learning Rate & [1e-5, 1e-3] \\
& Warmup Steps & [500, 1650] \\
& Weight Decay & [0.005, 0.15] \\
& Dropout Rate & [0.15, 0.3] \\
& $\lambda$ (Loss Weight) & [(0.1, 0.9),(0.4,0.6)] \\
& Training Epochs & [50, 150] \\
& Early Stopping Patience & [5, 20] \\
\midrule
& $\tau$ (RTE Parameter) & [5000, 20000] \\
& Time Granularity & [5m, 1h] \\
& Max Prediction Window & [7d, 28d] \\
\midrule
& $\beta_1$ & [0.85, 0.95] \\
& $\beta_2$ & [0.995, 0.9999] \\
& $\varepsilon$ & [1e-9, 1e-7] \\
& Gradient Clipping & [0.5, 5.0] \\
\bottomrule
\bottomrule
\end{tabular}
\end{table}


\section{Mathematical Notations and Definitions}
\label{sec:appendix_notation}

\cref{tab:notations} summarizes the notations used throughout the paper to describe social outbreak events, associated posts, and the engagement predictions. Each notation is accompanied by a brief explanation for clarity and ease of reference.
\begin{table}[htb]
\caption{Notation Table}
\label{tab:notations}
\begin{adjustbox}{max width=1.0\linewidth}
\begin{tabular}{ll}
\toprule
\toprule
\textbf{Notation} & \textbf{Description} \\
\midrule
$\mathcal{E}$ & A social outbreak event \\
$\mathcal{P}$ & The set of associated posts for $\mathcal{E}$, $\mathcal{P} = \{p_1, p_2, \dots, p_N\}$ \\
$p$ & A single post $p \in \mathcal{P}$ \\
$t_0$ & The original posting time of a post $p$ \\
$x$ & The textual content of a post $p$ \\
$u$ & User metadata associated with the post \\
$\mathcal{O}$ & The set of possible opinion classes \\
$o$ & The opinion class of a post, $o \in \mathcal{O}$ \\
$\mathcal{P}_o$ & The set of posts sharing opinion $o$, $\{p \in \mathcal{P} \mid o(p) = o\}$ \\
$H$ & Interval-censored engagement history of a post, \\ & $H = \{(t_1, e_1), \dots, (t_m, e_m)\}$ \\
$t_j$ & Observation time for engagement measurement in $H$ \\
$e_j$ & $d$-dimensional vector capturing different types of \\ & engagement at time $t_j$ \\
$\Delta t_j$ & Interval between consecutive observations, \\ & $\Delta t_j = t_{j+1} - t_j$ \\
$\tau_{obs}$ & Observation window duration (e.g., 1 day) \\
$H_{\tau_{obs}}(p)$ & Initial engagement history within observation window, \\ & $\{(t, e) \in H \mid t_0 \leq t \leq t_0 + \tau_{obs}\}$ \\
$T$ & Prediction horizon (e.g., 28 days) \\
$\tau_{\text{step}}$ & Fixed time interval for predictions (e.g., 5 minutes) \\
$K$ & Number of prediction points, $K = \lfloor T / \tau_{\text{step}} \rfloor$ \\
$\hat{e}(t)$ & Predicted engagement at time $t$ \\
$\hat{e}_{\text{total}}$ & Predicted total cumulative engagement over horizon $T$ \\
\bottomrule
\bottomrule
\end{tabular}
\end{adjustbox}
\end{table}

\section{Next-time Social Engagement Prediction}
\label{sec:next_token_pred}
While overall engagement prediction provides valuable insights into model performance, the ability to predict engagement at future time points is crucial for real-time social media monitoring and intervention. This section focuses on models capable of generating predictions for upcoming engagement values at the next future time point. We evaluate Informer~\cite{zhou2021informer}, Autoformer~\cite{wu2021autoformer}, Mamba~\cite{mamba2}, and \icmamba for this task, as these models are architecturally designed for next-point prediction.

We follow the same temporal set up for next-time social engagement prediction task, with 6-hour history data as the input.
For this task, we set up three different temporal stages of post lifecycle, each representing different length of intervals and data availability scenarios. 
Early-stage predictions (within first hour) have limited historical data but require quick response to emerging trends. 
Mid-stage predictions (within first day) balance data availability with evolving engagement patterns. 
Late-stage predictions (within first week) have rich historical context but must account for long-term engagement dynamics.

\begin{table}[htb]
    \caption{Engagement prediction results with fixed 6-hour historical window; RMSE scores reported; lower is better; best results in boldface. All models use exactly 6 hours of historical data regardless of when the next engagement occurs.}
    \centering
    \begin{adjustbox}{max width=1.0\linewidth}
    \begin{tabular}{lccccc}
    \toprule
    \toprule
    Model                              & Bushfire       & Climate        & Vaccination    & CoVID          & DiN            \\ \midrule
    Informer~\cite{zhou2021informer}   & 0.208          & 0.215          & 0.203          & 0.211          & 0.248          \\
    Autoformer~\cite{wu2021autoformer} & 0.196          & 0.203          & 0.191          & 0.199          & 0.234          \\
    \midrule
    Mamba~\cite{mamba2}                & 0.152          & 0.158          & 0.147          & 0.155          & 0.184          \\
    \icmamba                           & \textbf{0.144} & \textbf{0.150} & \textbf{0.139} & \textbf{0.147} & \textbf{0.175} \\
    \bottomrule
    \bottomrule
    \end{tabular}
    \end{adjustbox}
    \label{tab:fixed_window}
\end{table}

\begin{table}[htb]
\caption{Early-stage engagement prediction results (next interval $\leq$ 1 hour); RMSE scores reported; lower is better; best results in boldface.}
\centering
\begin{adjustbox}{max width=1.0\linewidth}
\begin{tabular}{lccccc}
\toprule
\toprule
Model                              & Bushfire       & Climate        & Vaccination    & CoVID          & DiN            \\ \midrule
Informer~\cite{zhou2021informer}   & 0.245          & 0.252          & 0.238          & 0.248          & 0.299          \\
Autoformer~\cite{wu2021autoformer} & 0.231          & 0.238          & 0.225          & 0.234          & 0.284          \\
\midrule
Mamba~\cite{mamba2}                & 0.188          & 0.194          & 0.193          & 0.201          & 0.265          \\
\icmamba                           & \textbf{0.169} & \textbf{0.175} & \textbf{0.164} & \textbf{0.172} & \textbf{0.235} \\
\bottomrule
\bottomrule
\end{tabular}
\end{adjustbox}
\label{tab:early_stage}
\end{table}

\begin{table}[htb]
\caption{Mid-stage engagement prediction results (next interval $\leq$24 hours); RMSE scores reported; lower is better; best results in boldface.}
\centering
\begin{adjustbox}{max width=1.0\linewidth}
\begin{tabular}{lccccc}
\toprule
\toprule
Model                              & Bushfire       & Climate        & Vaccination    & CoVID          & DiN            \\ \midrule
Informer~\cite{zhou2021informer}   & 0.198          & 0.205          & 0.193          & 0.201          & 0.235          \\
Autoformer~\cite{wu2021autoformer} & 0.187          & 0.194          & 0.182          & 0.190          & 0.223          \\
\midrule
Mamba~\cite{mamba2}                & 0.142          & 0.148          & 0.137          & 0.145          & 0.172          \\
\icmamba                           & \textbf{0.135} & \textbf{0.141} & \textbf{0.130} & \textbf{0.138} & \textbf{0.164} \\
\bottomrule
\bottomrule
\end{tabular}
\end{adjustbox}
\label{tab:mid_stage}
\end{table}

\begin{table}[htb]
\caption{Late-stage engagement prediction results (next interval $\leq$1 week); RMSE scores reported; lower is better; best results in boldface.}
\centering
\begin{adjustbox}{max width=1.0\linewidth}
\begin{tabular}{lccccc}
\toprule
\toprule
Model                              & Bushfire       & Climate        & Vaccination    & CoVID          & DiN            \\ \midrule
Informer~\cite{zhou2021informer}   & 0.165          & 0.137          & 0.152          & 0.168          & 0.175          \\
Autoformer~\cite{wu2021autoformer} & 0.156          & 0.128          & 0.163          & 0.159          & 0.168          \\
\midrule
Mamba~\cite{mamba2}                & 0.118          & 0.123          & 0.114          & 0.121          & 0.144          \\
\icmamba                           & \textbf{0.112} & \textbf{0.117} & \textbf{0.108} & \textbf{0.115} & \textbf{0.137} \\
\bottomrule
\bottomrule
\end{tabular}
\end{adjustbox}
\label{tab:late_stage}
\end{table}

\paragraph{Fixed-Window Prediction (6-Hour Input)} Table~\ref{tab:fixed_window} presents the RMSE scores for engagement prediction using a fixed 6-hour historical window. All models use exactly 6 hours of historical data regardless of when the next engagement occurs. 
We observe that \icmamba consistently achieves the lowest RMSE scores across all datasets, indicating strong performance in capturing short-term temporal patterns. However, it's noteworthy that baseline models like Autoformer and Informer also perform competitively, suggesting that the fixed-window approach provides sufficient context for short-term prediction.
An interesting finding is that the performance gap between \icmamba and Mamba is relatively small in this setting.

\paragraph{Early-Stage Prediction ($\leq$ 1 hour)}
In the early-stage prediction task, models forecast the next engagement within the first hour of a post's publication. As shown in Table~\ref{tab:early_stage}, all models experience increased RMSE compared to the fixed-window prediction, reflecting the challenge of making accurate predictions with limited historical data (typically 2-9 data points). Notably, \icmamba achieves the lowest RMSE, but the performance gap between \icmamba and Mamba widens in this setting.

Another observation is that the baseline models, Informer and Autoformer, show a heavy drop in performance during early-stage predictions. 
Additionally, the DiN dataset shows higher RMSE scores across all models, indicating that early-stage prediction is particularly challenging for post related to disinformation. 

\paragraph{Mid-Stage Prediction ($\leq$24 hours)}
In the mid-stage predictions, with more historical data available, all models show improved RMSE scores (Table~\ref{tab:mid_stage}). The performance gap between the models becomes smaller, indicating that the availability of additional data helps all models make better predictions. \icmamba continues to outperform the baselines.

An interesting finding is that the performance on Vaccination theme shows a significant reduction in RMSE for all models in the mid-stage prediction. This may imply that engagement patterns for vaccination-related content become more predictable within the first day, possibly due to sustained public interest and consistent interaction patterns.

\paragraph{Late-Stage Prediction ($\leq$1 week)}
 For late-stage predictions, with extensive historical data (up to one week), all models achieve their best RMSE scores (Table~\ref{tab:late_stage}). The performance differences between models are less pronounced, though \icmamba still holds a slight advantage.

 We found that the Climate theme shows relatively low RMSE scores across all models in late-stage prediction. This could reflect consistent engagement patterns over longer periods for climate-related content, perhaps due to sustained public interest and ongoing discussions.

\end{document}